\documentclass[
]{ceurart}

\usepackage{commath}
\usepackage{graphics}
\usepackage{siunitx}
\usepackage{subfig}

\begin{document}

\copyrightyear{2023}
\copyrightclause{Copyright for this paper by its authors.
  Use permitted under Creative Commons License Attribution 4.0
  International (CC BY 4.0).}

\conference{De-Factify: Workshop on Multimodal Fact-Checking and Hate Speech Detection, co-located with AAAI 2023. 2023
Washington DC, USA}

\title{Logically at Factify 2: A Multi-Modal Fact Checking System Based on Evidence Retrieval techniques and Transformer Encoder Architecture}
\author{Pim Jordi Verschuuren}[
email=Pim.jv@logically.co.uk
]

\author{Jie Gao}[%
orcid=0000-0002-3610-8748,
email=jie@logically.co.uk
]

\author{Adelize van Eeden}[%
email=adelize.ve@logically.co.uk
]

\author{Stylianos Oikonomou}[
email=stylianos@logically.co.uk
]

\author{Anil Bandhakavi}[%
email=anil@logically.co.uk,
url=https://www.logically.ai/team/leadership/anil-bandhakavi,
]

\address{Brookfoot Mills, Brookfoot Industrial Estate, Brighouse, HD6 2RW, United Kingdom}


%

\begin{abstract}
In this paper, we present the Logically submissions to De-Factify 2 challenge (DE-FACTIFY 2023) on task 1 of Multi-Modal Fact Checking. We describe our submission to this challenge including explored evidence retrieval and selection techniques, pre-trained cross-modal and unimodal models, and a cross-modal veracity model based on the well established Transformer Encoder (TE) architecture which heavily relies on the concept of self-attention. Exploratory analysis is also conducted on the Factify 2 data set that uncovers the salient multi-modal patterns and hypothesis motivating the architecture proposed in this work. A series of preliminary experiments were done to investigate and benchmark different pre-trained embedding models, evidence retrieval settings and thresholds. The final system, a standard two-stage evidence based veracity detection system, yielded a weighted average F1 score of 0.79 on both the validation set and final blind test set of task 1, which achieved 3rd place with a small margin to the top performing systems on the leaderboard among 9 participants. 


\end{abstract}


\maketitle

\section{Introduction}

Misinformation and fake news can spread rapidly and cause harm at various levels. One way to protect ourselves from these negative impacts is through fact-checking and debunking false information with evidence-based reporting. However, this process can be resource-intensive and time-consuming. To address this issue, researchers have developed automated fact-checking systems using deep learning techniques, which can handle tasks such as claim detection, claim matching, evidence retrieval, and veracity prediction using natural language processing techniques on textual content. While there has been progress in this area, there is still a need for multimodal approaches that can handle both text and image inputs. To address this gap, this paper presents a multimodal veracity prediction system for automated fact-checking which was developed as part of the Factify 2 competition organized by De-Factify@AAAI 2023.

The remainder of the paper is structured as follows: Section 2 presents a brief overview of related work and section 3 describes our general framework and model architecture. Section 4 discusses the dataset supplied by the Factify 2 competition followed by an overview of our experiments in section 5. Section 6 and 7 present the final results and our conclusions, respectively.

\section{Related Work}\label{sec:rel}
As an essential part of automated fact verification, effective techniques for modeling claim-evidence for veracity prediction have been a hot topic and key research questions in existing fact-checking methods. Most of the recent work focuses on using textual evidence in veracity prediction of which there are mainly two lines of work. One direction \cite{wadden2022multivers,stammbach2021evidence,gao2021log} is to use a single document (such as is provided in the Factify task dataset) with long text evidence and through leveraging models constructed for long sequences. Examples such as BigBird \cite{zaheer2020big}, Longformer\cite{beltagy2020longformer} and recent advancements in the ConvNets architecture witnessed in the Long Range Arena leaderboard (e.g., Mega \cite{ma2022mega}, S5\cite{smith2022simplified}) are seen to obtain top results in a wide range of tasks and other leaderboards. The benefits of exploiting long-sequence models at document level is a) the simplicity of the overall architecture; b) allows to accommodate for more context of the whole article into modeling and natural language inference. An optimal setup of the maximum length for both claim (or query) and document sequence, and the document level veracity labels is commonly required \cite{zhao2021ror,wadden2022multivers,gao2021log}. The advantage of incorporating lots of context into inference is also seen in modeling question answering (QA) tasks \cite{zaheer2020big,beltagy2020longformer}, for which the document-level veracity labels are relatively "cheap" to obtain. The downside of using a simple long-text model technique at document-level is the lack of interpretability (w.r.t. evidence selection), it is computational expensive, the limitation in dealing with the complexity of certain (multi-hop) claims \cite{nakano2021webgpt}, and lack of diversity and scalability when dealing with a large amount of diverse documents in a real-world application. These constraints were more apparent in open domain fact checking tasks that make use of web data extracted with commercial search engines as building blocks in fact-checking systems in order to incorporate more diverse sources. It is worth to note that long-sequence models can be adapted for the purpose of evidence selection e.g., through framing the task as a token-level prediction task. For instance, as one of the top systems in the SciFact leaderboard \footnote{https://leaderboard.allenai.org/scifact/submissions/public}, LongChecker \cite{wadden2020fact} used LongFormer \cite{beltagy2020longformer} for scientific claim verification with paragraph-level evidence selection. In their method, every sentences is inserted with a [CLS] token with global attention, which allows the model to predict on this sentence-level token as evidence. Most of these works focus on a limited context such as a few Wikipedia documents, a single article and abstracts or text snippets from either research literature or a small synthetic corpus.

Another line of work widely adopted and one of the key tasks in FEVER \cite{thorne-etal-2018-fever,aly2021feverous} is to involve evidence retrieval and selection. The framework exploits larger document context to extract evidentiary (or rationales) passages as first step and veracity prediction is then modeled to condition on the claim and the selected rationales. The evidentiary passages report the findings to the claim which can be used to justify each veracity label and can be selected at either sentence- or paragraph-level. Despite the revolutionary breakthroughs with Large-Scale Language Models (LSLMs), such as GPT-3\cite{floridi2020gpt} and ChatGPT\footnote{https://openai.com/blog/chatgpt/}, and their impressive generative capabilities, these large models are still lacking key zero-shot or few-shot learning capabilities needed for fact checking tasks. This is mainly due to their incorrectly retrieved, incomplete or outdated knowledge stored in their weights which makes these techniques susceptible to hallucinations \cite{maynez2020faithfulness,lazaridou2022internet}, which is conflicting with fact checking tasks that require factuality as an essential element in modeling. Moreover, an efficient approach to keep LSLMs up-to-date and grounded to ever-growing factual and new information is imperative but still unresolved to date. Recent work \cite{lazaridou2022internet,wei2021finetuned} shows that lightweight methods with fine-tuned and smaller models outperform these big models in a range of knowledge-intensive NLP tasks including Natural Language Inference (NLI), Recognizing textual entailment (RTE), Reading Comprehension (RC), QA, etc. Sentence-BERT (SBERT) \cite{reimers2019sentence} is one of the most popular techniques based on the BERT language model \cite{devlin2018bert} used for evidence selection \cite{saakyan2021covid,yao2022end} which can be framed as a sentence-pair regression task. SBERT models are used to encode contextualized representations for each of the evidence passages which are then ranked according to their semantic similarity with the contextualized representation of the corresponding claim. In the final step, top $k$ evidentiary passages are selected for veracity prediction. The challenge of this multi-staged verification framework is 1) the rationales extracted out-of-context may lack information required to make a prediction (e.g., acronyms, unresolved coreferences); 2) the evidence extraction (through passages ranking) requires high quality training data that is costly to obtain with domain experts from both closed and open domain tasks \cite{barron2020overview}. Various efforts to address the constraints have been undertaken to explore 1) paragraph level train data from scientific literature with paper title as claim and abstract as evidence as high-precision heuristics (e.g., SciFact \cite{wadden2022multivers}); 2) QA dataset with question and answer considered as claim and evidence respectively \cite{lee2018improving}; 3) NLI dataset with the claim as hypothesis and evidence as evidence \cite{schuster2021get}. We follow a second line of work for which the evidence retrieval component is implemented in our system following current SoTA methods.

Automated multi- or cross-modal fact checking is an underdeveloped field compared to text-based techniques. Recent developments have shown that cross-modal pre-trained models (e.g.,VideoBERT \cite{sun2019video}, VisualBERT \cite{li2019visual}, Uniter \cite{chen2019uniter}, CLIP \cite{radford2021learning}) have achieved significant results in downstream cross-modal tasks \cite{gao2021clip,guo2022calip,jiang2022comclip} with great transferability for zero-shot or few-shot scenarios. Our work is inspired by \cite{yao2022fact}, which was one of the initial explorations in multimodal fact-checking task. In their proposed method, the Contrastive Language–Image Pre-training (CLIP) model \cite{radford2021learning}) is adopted as encoder to learn joint language-image embedding between each image and input claim text. Top-5 candidate image evidences are taken as input along with multimodal claim for multimodal claim verification model with a simple cross-attention network. It is worth noting that the CLIP model allows to model image-text contextual alignment at coarse-grained contextual (global) level but ignores the compositional matching of disentangled concepts (i.e., finer-grained cross-modal alignment at region-word level)\cite{jiang2022comclip,messina2021fine,messina2021fine}.

\section{Methodology} \label{sec:method}
\subsection{Problem statement}
We frame the Factify 2 problem as a multimodal entailment task as in the previous submission \cite{gao2021log}, which considers a multimodal claim $c = c_{text} + c_{image}$ as hypothesis and a multimodal document $d = d_{text} + d_{image}$ as premise. The goal is to learn a function $f(c,d)$ that infers one of the five entailment categories including "Support\_Multimodal", "Support\_Text", "Refutes", "Insufficient\_Multimodal" and "Insufficient\_Text". Additional details on the task can be found in \cite{surya2023factify2}.



\subsection{General Architecture}
Our system architecture follows a standard two-stage claim verification approach as established through various shared tasks in recent years, typically FEVER\cite{thorne2018fact}, FEVER 2.0 \cite{thorne2019fever2}, FEVEROUS \cite{aly2021fact} and SCIVER \cite{wadden2021overview}. First, a textual evidence retrieval component identifies from a given document the evidence passages most relevant to the corresponding claim text. Then, a transformer based cross-modal model is trained on all the input across multimodalities including selected evidence passages text, claim text, claim image, document image, claim OCR text and document OCR text to predict five multimodal entailment categories with respect to the multimodal claim. A pre-trained cross-modal model (i.e. CLIP) and a pre-trained text embedding model are both employed in the embedding layer in order to learn a cross-modal matching model using both unified-multimodal and unimodal representations. Overall, the implemented architecture adopts a list-wise concatenation strategy \cite{jiang2021exploring} which is one of common strategies in most recent sequence-to-sequence SoTA veracity prediction models.



\begin{figure}[h]
\centering
\includegraphics[width=0.90\linewidth]{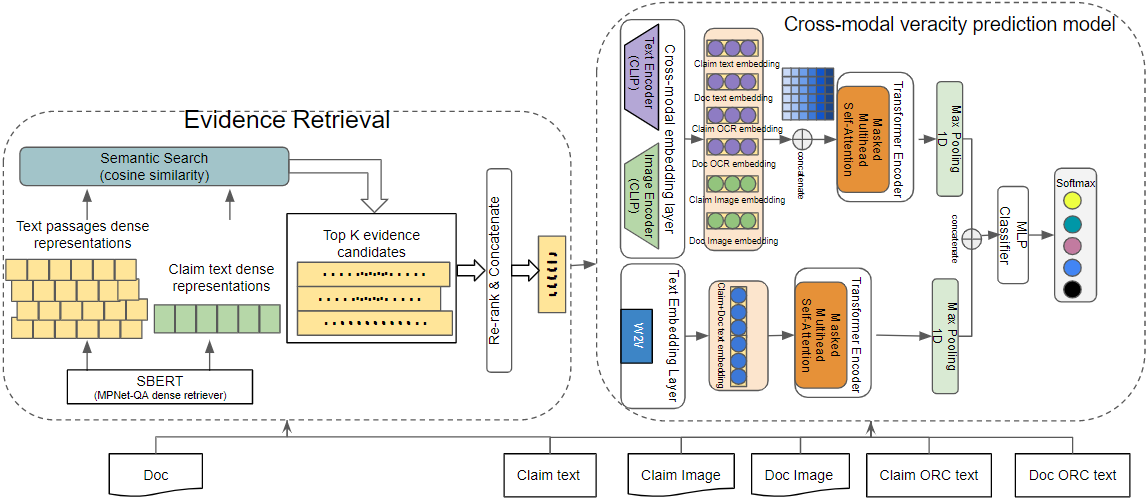}
\caption{Logically General System Architecture}
\label{fig:system_architecture}
\end{figure}

\subsection{Evidence Retrieval}\label{sec:method_evidence_retrieval}
In evidence retrieval, `multi-qa-mpnet-base-dot-v1` \footnote{The model is available on the Hugging Face hub, accessible via https://huggingface.co/sentence-transformers/multi-qa-mpnet-base-dot-v1} and is employed to compute embeddings for both claim text and document text at passage level. In terms of passage granularity, both paragraph- and sentence-level retrieval have been experimented with (see Section \ref{sec:experiment}). This is a SBERT model based on the MPNet architecture \cite{song2020mpnet} and is trained on a Question-Answer (QA) dataset with 215M QA pairs from diverse sources. The model was tuned for a semantic search using a dot-product score function in order to find relevant passages corresponding to a given query. The model encodes text into a 768-d vector and supports 512 maximum number of tokens.

Regarding the similarity computation and semantic search, we use a simple dot product with the normalised SBERT embeddings (as proxy to cosine similarity) which enables a quick and efficient passage ranking and scalability of up to about 1 Million entries.

Top $K$ passages obtained from the semantic search are then re-ranked based on their relevancy to the claim text and concatenated into a longer text snippet before being fed into the cross-modal veracity prediction model.

\subsection{Embedding Layer}\label{sec:method_embedding_layer}

Our embedding layer consists of a cross-modal encoder and a unimodal text encoder. We hypothesize that modeling solely on text-to-text interaction (i.e., text premise and hypothesis) can supplement the modeling solely on cross-modal premise and hypothesis interaction and vice versa. This architecture facilitates the measuring of multimodal semantic relatedness in this multimodal fact checking task by mapping more textual alignment signals into subsequent semantic space. This considers that text specific models can capture more accurate and semantically meaningful word- or sentence-level alignment. 

The cross-modal encoder is implemented with a pre-trained CLIP model that aims to map visual and text embeddings into a common space. The ViT-B/32 variant (ViT-Base with patch size 32) is chosen in this work because of its smaller amount of parameters, less FLOPS and greater inference speed. ViT-B/32 consists of a text encoder and an image encoder which are used to encode text inputs (including claim text, evidentiary passage and two images OCR text) and image inputs (including claim image and document image) respectively before concatenating into a $6 \times 512$ matrix as a single input to the subsequent transformer encoder. The CLIP architecture allows for a maximum input text length of 77 tokens. The pre-trained Word2vec model ("Word2vec Google News 300") \cite{mikolov2013distributed} is adopted as a unimodal text encoder. It encodes the concatenated text sequence of claim and document evidentiary passage text, and obtains a 300-D feature vector for each token. Zero-padding is applied to match the longest sentence in the training set. Both the pre-trained CLIP and Word2Vec embedding model were not fine-tuned.




\subsection{Cross-modal veracity prediction}\label{sec:method_multimodal_entailment}
The second component of veracity prediction is based on the well established Transformer Encoder (TE) architecture, which heavily relies on the concept of self-attention \cite{vaswani2017att} to effectively model higher-order interactions and context in an input. Recent research has shown that multi-head self-attention mechanisms and transformer architectures are computationally efficient and accurate in this regard. The self-attention mechanisms of the TE encoder allows for simple but powerful reasoning that can identify hidden relationships between vector entities, regardless of whether they are visual or textual in nature. Therefore, our cross-modal veracity prediction model is implemented based on self-attention mechanisms to learn the joint distribution of text representations of claim-document text pair and cross-modal feature representations of all modalities contained in claim and document. 

Specifically, the claim and document embeddings of joint input by CLIP and text input by text embedding layer are passed through two separate transformer encoders \cite{vaswani2017att} consisting of $N$ identical sequential blocks of a multi-head attention (MHA) and a fully connected feed-forward network (FFN). Within each transformer encoder, multiple blocks allows for a deeper understanding of the inputs. For each block the input $x$ is passed through a multi-head attention layer of which the output is added to the initial input such that information in the initial sequence is not lost. Layer normalization is applied to the output to allow for faster training and small regularization i.e. $x = \mbox{LayerNorm}(x + \mbox{MHA}(x))$. The output is then passed to a feed-forward network to allow for more model complexity. The output is again added to the original input and layer normalization is applied i.e. $x = \mbox{LayerNorm}(x + \mbox{FFN}(x))$. The output of the final block (i.e., the output of each transformer encoder in the diagram) is passed through an adaptive max pooling layer to reduce the output dimensions. The output of two separate transformer encoders are then concatenated before feeding into a MLP classifier for the five category prediction. The five categories probabilities are obtained from the final output softmax layer.




\section{Factify Dataset}\label{sec:data}



\subsection{Dataset Description}

The Factify 2 dataset created and supplied by the organisers covers a train, validation, and test set. The train set contains 35000 data pairs, while the validation and test sets each contain 7500 data pairs. Each data pair consists of a claim and a document, each of which comprises an image, a text, and an OCR text extracted from the image. The data pairs are annotated with one label from 5 categories including Support\_Multimodal, Support\_Text, Refute, Insufficient\_Multimodal, or Insufficient\_Text.

\subsection{Text Length Distribution}

\begin{figure}[ht]
  \centering
  \subfloat[Claim Text Length]{\includegraphics[width=0.47\textwidth]{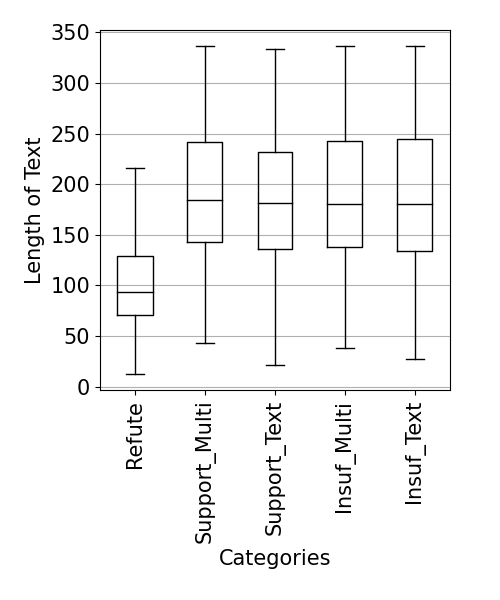}}
  \hfill
  \subfloat[Document Text Length]{\includegraphics[width=0.47\textwidth]{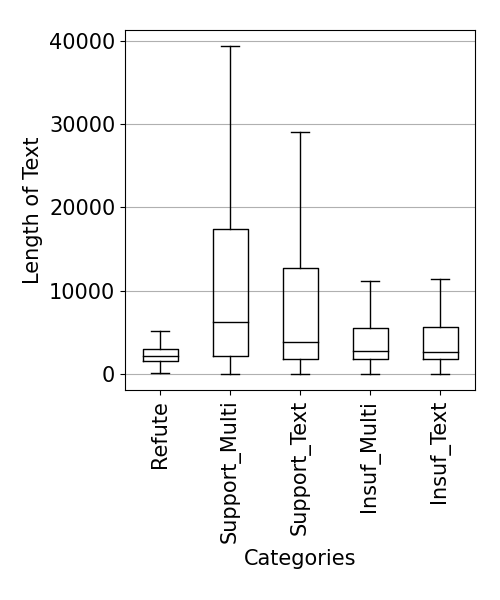}}
  
  \caption{Boxplot of Text Length Distribution of all Categories}\label{fig:text_len}
\end{figure}

\begin{figure}[ht]
  \centering
  \subfloat[Claim OCR Text Length]{\includegraphics[width=0.47\textwidth]{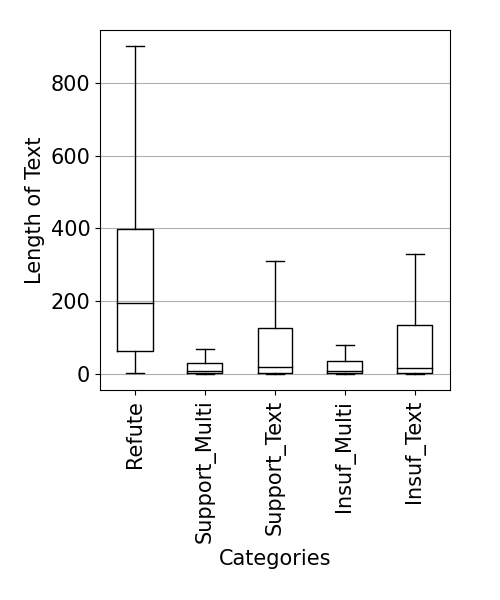}}
  \hfill
  \subfloat[Document OCR Text Length]{\includegraphics[width=0.47\textwidth]{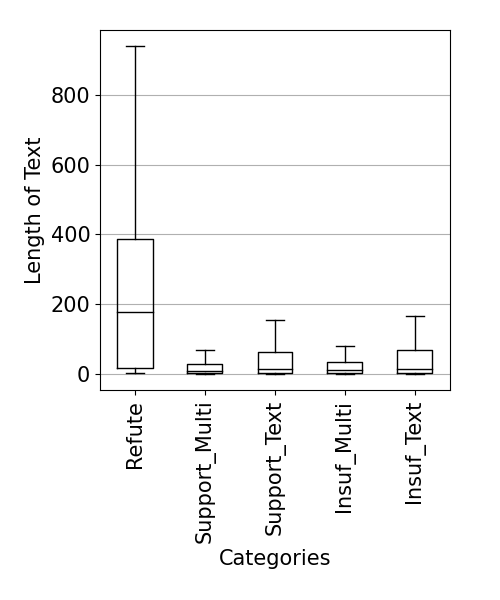}}
  
  \caption{Boxplot of OCR Text Length Distribution of all Categories}\label{fig:ocr_text_len}
\end{figure}

The training set text and OCR text length distributions are represented in Figures \ref{fig:text_len} and \ref{fig:ocr_text_len}. The text length distribution varies between the claim and document text, with the document text that tends to be much longer. This is expected as it is used to verify the claim. From Figure \ref{fig:text_len} (a), we can can see that claim text is much shorter and less varied for the Refute category than for the rest of the categories, which all have similar claim text length distributions. Figure \ref{fig:text_len} (b) shows that the Support\_Multimodal and Support\_Text categories have the larger spread of document text lengths and also the longest document text lengths. The two Insufficient categories have on average a smaller document text length, and Refute has the smallest variance and maximum length in document text length.

Considering the claim OCR length we see from Figure \ref{fig:ocr_text_len} that the Refute category has a much larger claim OCR length distribution and maximum length than any other category. The second largest claim OCR length distributions are the Support\_Text and the Insufficient\_Text categories, which then leaves the two Multimodal categories with the shortest claim OCR text lengths. The document OCR length distribution is very similar to that of the claim OCR, from Figure \ref{fig:ocr_text_len}b we see the only real difference is that the two Text categories have a smaller document OCR length distrubution than that of the claim OCR.

\subsection{Image Similarity Distribution}

\begin{figure}[h]
  \centering
  
  \subfloat[Claim Image and Document Image Similarity Score Histogram]{\includegraphics[width=0.85\textwidth]{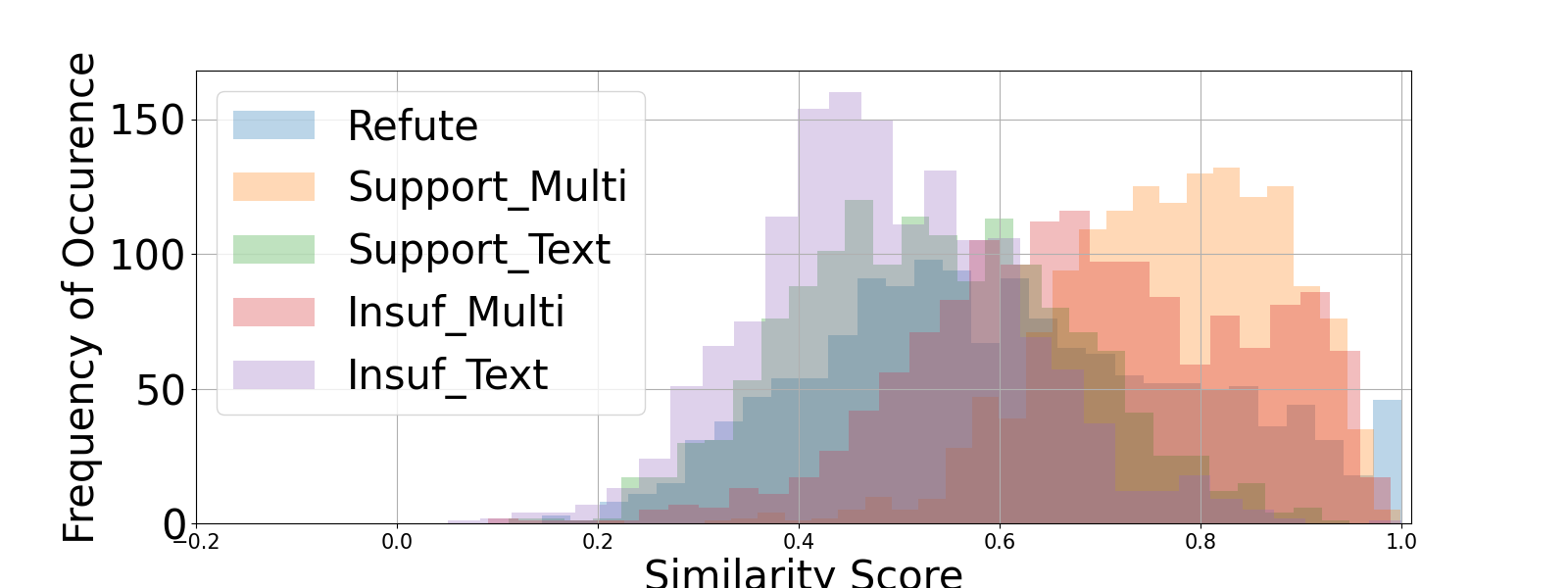}\label{fig:claim_img_doc_img_simscore_hist}}\\

  \subfloat[Claim Image and Document Image Similarity Boxplot]{\includegraphics[width=0.5\textwidth]{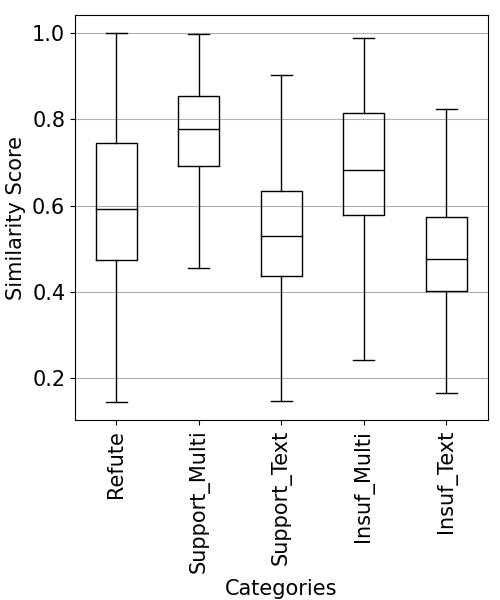}\label{fig:claim_img_doc_img_simscore_box}}
  
  \caption{Claim Image and Document Image Similarity Distribution}
\end{figure}

An image similarity investigation was conducted in order to gain an intuition of the similarity between the claim and document images for each category. Using image pairwise CLIP embeddings we calculate a similarity score and analyse it per category. Figures \ref{fig:claim_img_doc_img_simscore_hist} and \ref{fig:claim_img_doc_img_simscore_box} illustrate that the similarity between the claim and document image is comparatively higher within the categories for Support\_Multimodal and Insufficient\_Multimodal than the other categories. The label correlation with similarity of image pairs has largely increased compared to factity 1 dataset \cite{gao2021log} of last year.  This further indicates that there is explicit correlation within the multimodal categories which can be leveraged to learn and verify multimodal entailment categories.

\subsection{Multimodal Similarity Distribution}

\begin{figure}[h]
  \centering
  
  \subfloat[Claim Text and Document Image Similarity Score Histogram]{\includegraphics[width=0.85\textwidth]{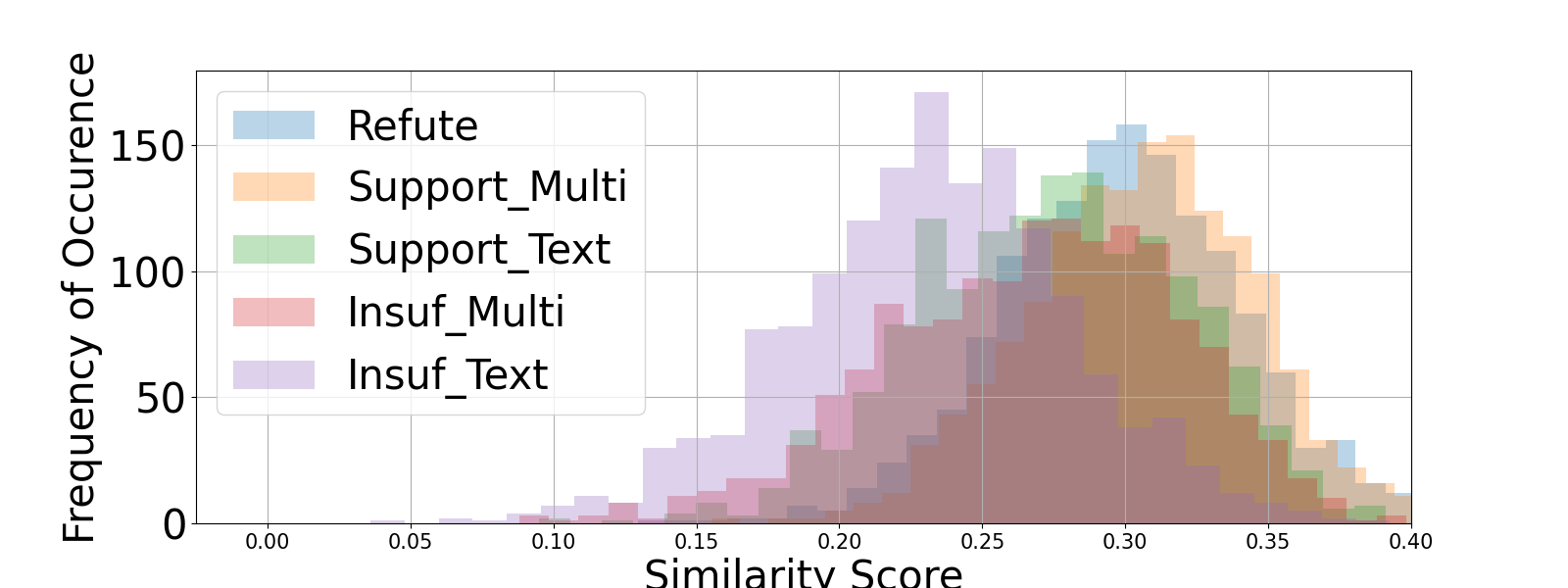}\label{fig:claim_txt_doc_img_simscore_hist}}\\

  \subfloat[Claim Text and Document Image Similarity Boxplot]{\includegraphics[width=0.5\textwidth]{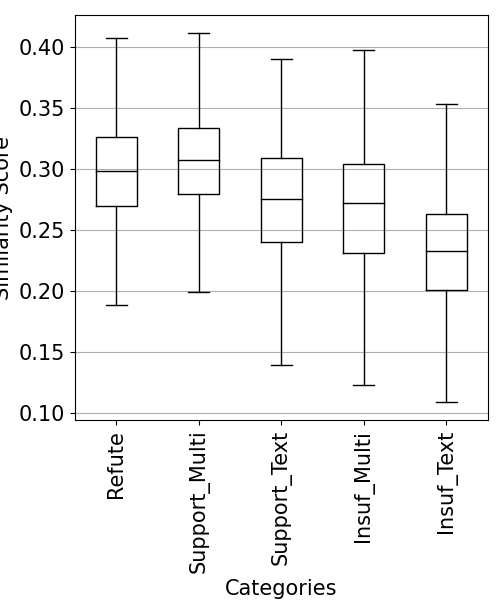}\label{fig:claim_txt_doc_img_simscore_box}}
  
  \caption{Claim Text and Document Image Similarity Scores}
\end{figure}

The multimodal CLIP similarity among multimodal claim and doc pairs is explored to investigate our hypothesis that doc image should contain content that is related to the claim in order to entail either support or refute verdict decisions. Figures \ref{fig:claim_txt_doc_img_simscore_hist} and \ref{fig:claim_txt_doc_img_simscore_box} depict the cosine similarity scores between the claim text and document image. It is noticeable that “Support\_Multimodal” presents the highest pairwise similarity correlation between label and claim-evidence pair. “Insufficient text” have the lowest pairwise similarity correlation, although our initial hypothesis was that “Insufficient\_Multimodal” should have the lowest value. This analysis suggests that differentiating between the different categories based on the claim text and document image correlation could be challenging.


\begin{figure}[h]
  \centering
  \subfloat[Claim Image and Document Text Similarity Score Histogram]{\includegraphics[width=0.49\textwidth]{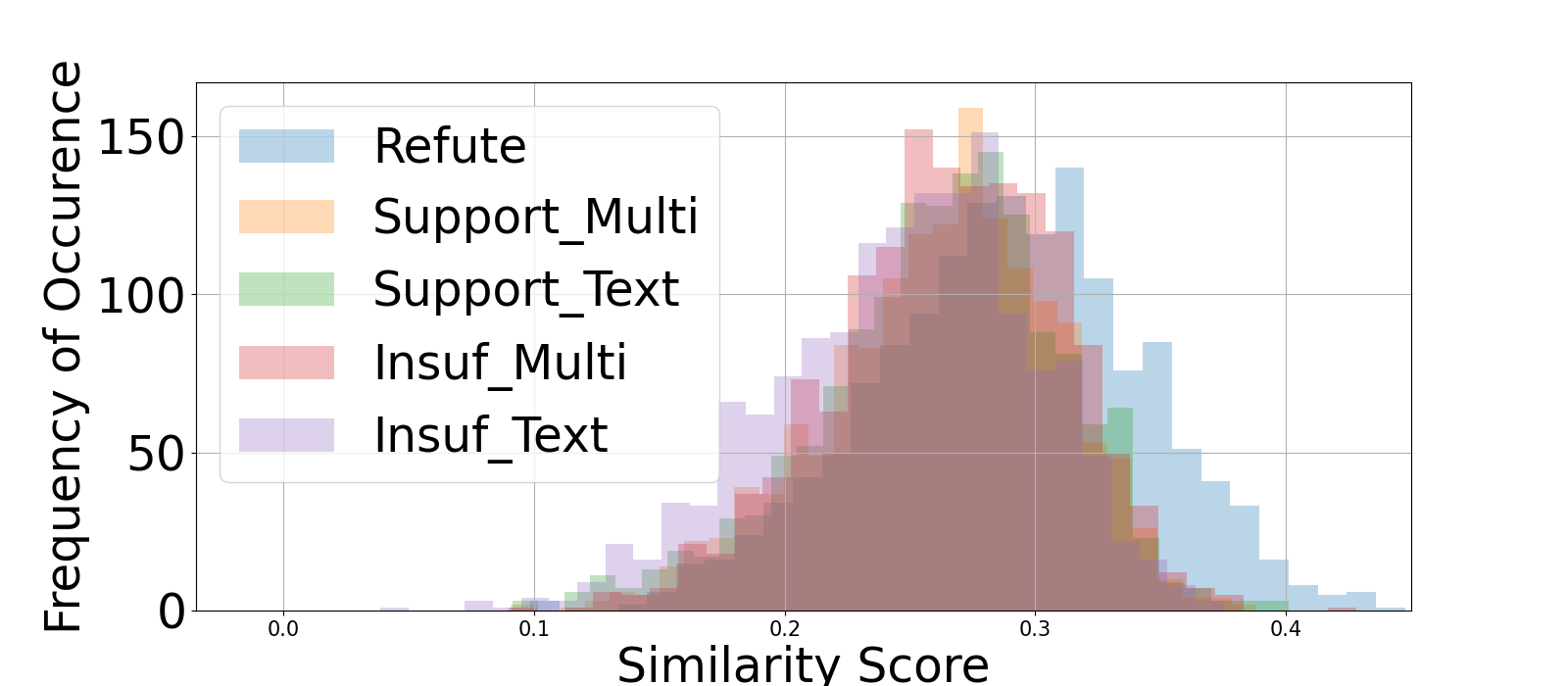}\label{fig:claim_img_doc_txt_simscore_hist}}
  \hfill
  \subfloat[Claim Image and Claim Text Similarity Score Histogram]{\includegraphics[width=0.49\textwidth]{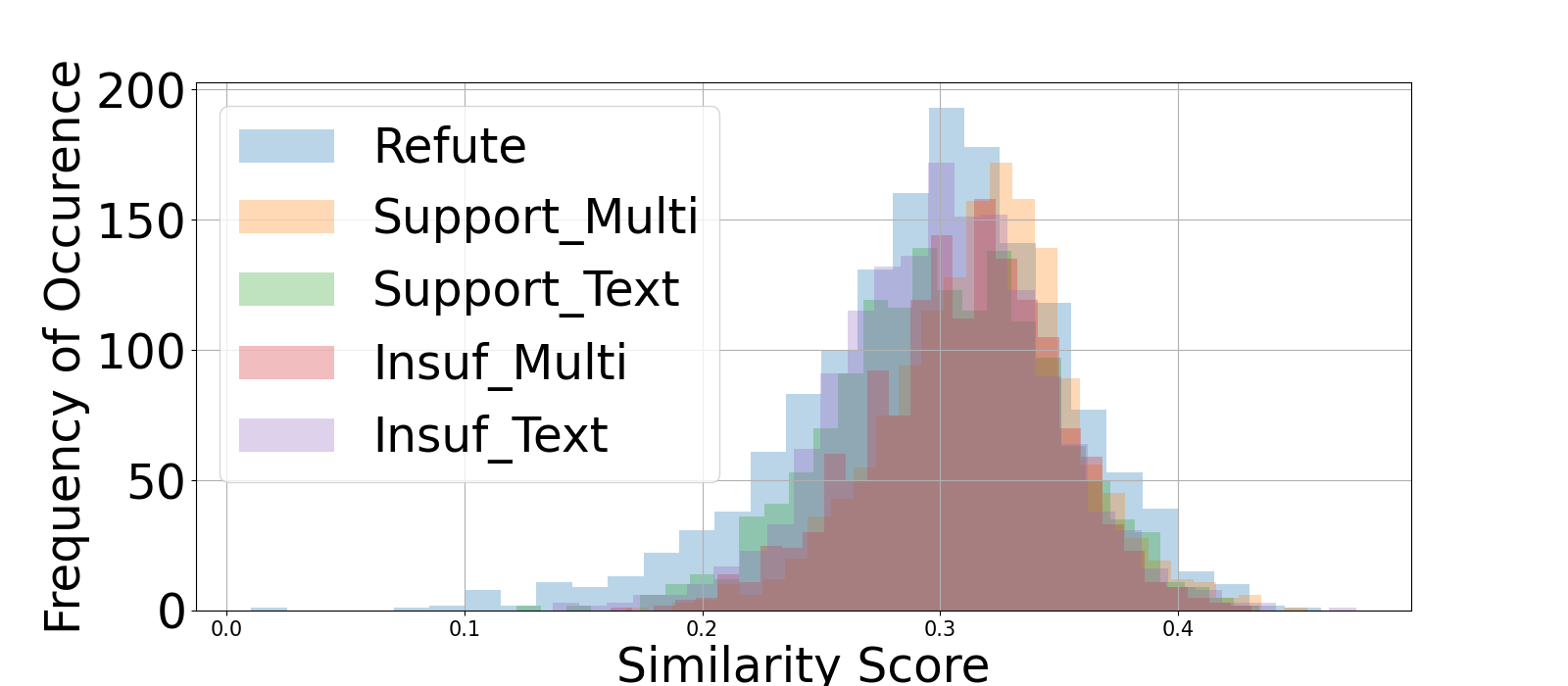}\label{fig:claim_img_claim_txt_simscore_hist}}\\

  \subfloat[Claim Image and Document Text Similarity Score Boxplot]{\includegraphics[width=0.475\textwidth]{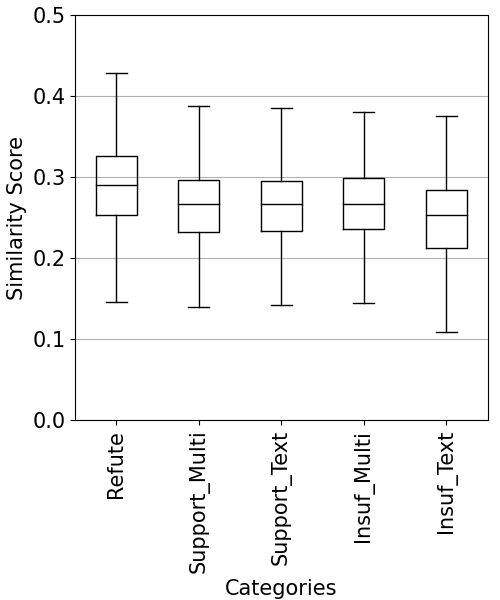}\label{fig:claim_img_doc_txt_simscore_box}}
  \hfill
  \subfloat[Claim Image and Claim Text Similarity Score Boxplot]{\includegraphics[width=0.475\textwidth]{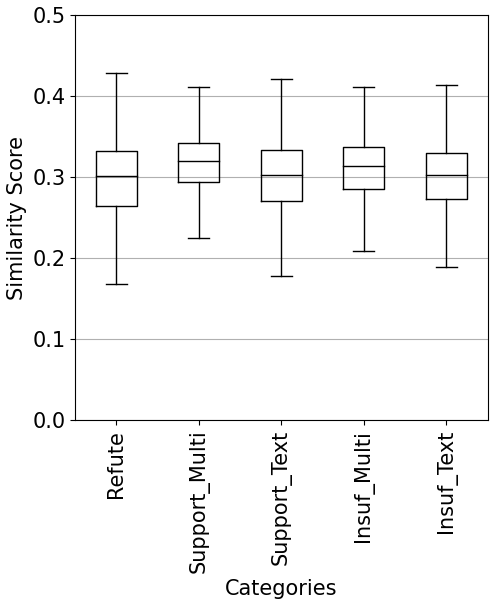}\label{fig:claim_img_claim_txt_simscore_box}}
  
  \caption{Image and Text Similarity distribution among multimodal claim and doc}
\end{figure}

In terms of correlation between the claim image and document text, due to the maximum text sequence constraints with CLIP, text access maximum length is truncated. Consequently, longer context of document text is not incorporated in this analysis. As shown in Figure \ref{fig:claim_img_doc_txt_simscore_hist} and \ref{fig:claim_img_doc_txt_simscore_box}, there is low degree of similarity correlation across the five categories, among which the "Refute" category shows highest similarity correlation. 

Lastly, Figure \ref{fig:claim_img_claim_txt_simscore_hist} and Figure \ref{fig:claim_img_claim_txt_simscore_box} show the similarity correlation between the claim image and the claim text, and show no significant deviation in similarity scores of different categories when the claim image and claim text are compared to each other. For the purpose of this task and this dataset, we hypothesize that the claim image should provide supplementary information to the claim text.

\section{Experiments}\label{sec:experiment}
\subsection{Model settings}



To validate and optimize the effect of evidence retrieval, we attempt to experiment with our model with 1) including or excluding evidence selection; 2) varying the length of evidence doc text sorted by evidence retriever; 3) passage ranking at paragraph level versus sentence level; 4) text-to-text alignment with SBERT versus cross-modal alignment with CLIP. Both SBERT and CLIP is used to rank evidence doc with paragraph and sentence level; 5) if SBERT model trained on QA dataset perform better than general purpose SBERT model. Note that ranking at paragraph level on top <5 or sentence level on top <5 is only an option for CLIP due to its maximum allowed length restriction.


For two transformer encoders, we choose an empirical setting of four heads in two MHAs. The number of sequential MHA and feed-forward network blocks per embedding input is $N_{blocks} = 2$. All our experiments are trained on a 3-layered MLP and the number of nodes per layer are set to 3072, 1024 and 5, respectively. A dropout of 0.5 and ReLU activations are applied between the MLP layers.

Preliminary experiments conducted in this work are elaborated in details as follows:
\begin{itemize}
    \item "model\_w/o\_ER": to validate the effectiveness with evidence retrieval, we remove evidence retrieval in our system and provide original document text to "Cross-modal veracity prediction model".
    \item "SBERT\_sentence\_ER\_top5":  One of the "top" \footnote{The best performing general purpose model is selected with a sorted list of model performances and recommended use cases provided by SBERT, accessible via https://www.sbert.net/docs/pretrained\_models.html} performing general purpose SBERT model ("all-MiniLM-L6-v2") is chosen in our experiment. This is an all-round model tuned for many use cases and 5 times faster while offering good quality compared to the best all-round model "all-mpnet-base-v2". The model is trained on a large and diverse dataset of over 1 billion training pairs and also fine-tuned for dot-product score function suitable for cosine similarity. The use of the all-round model allows us to evaluate the value of adopting QA fine-tuned counterpart that we hypothesize to be the optimal solution. Top 5 sorted sentences sorted by all-round SBERT model is configured in this setting.
    \item "SBERT\_sentence\_ER\_top10": Top 10 sorted sentence sorted by all-round SBERT model is configured in this setting.
    \item "SBERT\_sentence\_ER\_top15": Top 15 sorted sentence sorted by all-round SBERT model is configured in this setting.
    \item "SBERT-QA\_paragraph\_ER\_top5": SBERT QA dataset fine-tuned model (as described in \ref{sec:method_evidence_retrieval}) is adopted in this setting to obtain top 5 paragraphs as evidentiary passages for veracity inference in this setting.
    \item "SBERT-QA\_sentence\_ER\_top5": Top 5 sentences sorted by SBERT QA model and selected as evidentiary passages in this setting.
    \item "BigBird\_w/o\_ER": To evaluate the value of evidence selection against the long context modeling solution, the Google's BigBird pre-trained model fine-tuned on Factity dataset from last year \cite{gao2021log} is used to replace the Word2Vec model in the "Text Embedding layer" with this setting. This BigBird model allows a maximum 1396 tokens and contextual representation of text is adopted in this setting. 
\end{itemize}

\subsection{Training and validation}
For our experiment, the model was trained up to 80 epochs with early stopping on minimum validation loss by minimizing the cross-entropy loss function using the adaptive AdamW optimizer \cite{losh2017adamw} with an initial learning rate of $\gamma=1\mathrm{e}{-4}$ and epsilon $\epsilon=1\mathrm{e}{-8}$ with batch size $N_{batch}=16$. Early stopping patience is set to 5. A linear decreasing learning rate scheduler was used including $N_{steps}=438$ warming up training steps during which the learning rate increased linearly to the chosen learning rate. 

We have found that data scraping errors lead to invalid doc text content in the development dateset provided by organiser with 463 and 114 invalid samples in train and val set respectively. There also are 112 invalid samples in test set.  This results in document text containing only "We’ve detected that JavaScript is disabled in this browser ...". The invalid samples are removed from our training data.

\section{Results and Discussion}\label{sec:results}
The best model results in preliminary experiments described in section \ref{sec:experiment} are presented in Table \ref{tab:5wayresults_ex1}, Table \ref{tab:5wayresults_ex2} and Table \ref{tab:5wayresults_ex3} respectively. 

Firstly, the Table \ref{tab:5wayresults_ex1} shows that our veracity model without ER exhibits a reasonably good performance and utilising the long sequence model (BigBird) for text embeddings improves the base model with a small margin, by 1\% for all categories except "Refute". As comparison, further experiments with ER are conducted of which the results are presented in Table \ref{tab:5wayresults_ex2}  and Table \ref{tab:5wayresults_ex3}. The results in Table \ref{tab:5wayresults_ex2} indicate that all-round SBERT based evidence selection does not provide obvious performance improvement based on current preliminary explorations covering three top K sentences settings (K=5, 10, 15). In contrast, SERT-QA based model achieves big marginal improvement at both paragraph and sentence level. Our experiments covers both top 5 paragraphs and sentences, which improves best base model (without ER) by 1\% and 2\% respectively. Final results across 7 different experiment setup shows that combining SBERT-QA at top K sentence-level evidence passage retrieval achieves optimal performance compared to the base model without ER and the use of all-round SBERT model. The best model "SBERT-QA\_sentence\_ER\_top5" obtains 0.79 weighted average F1 at the 20th epoch.

\begin{table}
 \caption{5-way Classification Results of experiments without ER on val set}\label{tab:5wayresults_ex1}
 \small
 \resizebox{\columnwidth}{!}{
 \begin{tabular}{lSSSSSSSSSSSSS}
    \toprule
    \multirow{1}{*}{Categories} &
     \multicolumn{3}{|c}{model\_w/o\_ER} &
     \multicolumn{3}{|c}{BigBird\_w/o\_ER}  \\
     & {P} & {R} & {F1} & {P} & {R} & {F1} \\
     \midrule
    Support\_Multimodal & 0.73 & 0.79 & 0.76 & 0.73 & 0.81 & \textbf{0.77} \\
    Support\_Text  & 0.71 & 0.61 & 0.66 & 0.77 & 0.59 & \textbf{0.67} \\
    Insufficient\_Multimodal  & 0.66 & 0.66 & 0.66 & 0.64 & 0.70 & \textbf{0.67} \\
    Insufficient\_Text  & 0.71 & 0.75 & 0.73 & 0.73 & 0.75 & \textbf{0.74} \\
    Refute   & 0.99 & 0.98 & 0.98 & 0.98 & 0.98 & 0.98 \\
    \midrule
    Weighted Avg. & 0.76 & 0.76 & 0.76 & 0.77 & 0.77 & \textbf{0.77}\\
    \bottomrule
 \end{tabular}
 }
\end{table}

\begin{table}
 \caption{5-way Classification Results of experiments with all-round SBERT + ER on val set}\label{tab:5wayresults_ex2}
 \small
 \resizebox{\columnwidth}{!}{
 \begin{tabular}{lSSSSSSSSSSSSS}
    \toprule
    \multirow{1}{*}{Categories} &
     \multicolumn{3}{|c}{SBERT\_sentence\_ER\_top5} &
     \multicolumn{3}{|c}{SBERT\_sentence\_ER\_top10} & 
     \multicolumn{3}{|c}{SBERT\_sentence\_ER\_top15} \\
     & {P} & {R} & {F1} & {P} & {R} & {F1} & {P} & {R} & {F1} \\
     \midrule
    Support\_Multimodal & 0.72 & 0.85 & 0.78 & 0.74 & 0.78 & 0.76 & 0.75 & 0.77 & 0.76 \\
    Support\_Text  & 0.63 & 0.73 & 0.68 & 0.71 & 0.61 & 0.66 & 0.71 & 0.62 & 0.66 \\
    Insufficient\_Multimodal & 0.70 & 0.64 & 0.67 & 0.66 & 0.67 & 0.66 & 0.65 & 0.67 & 0.66 \\
    Insufficient\_Text  & 0.80 & 0.58 & 0.67 & 0.70 & 0.77 & 0.74 & 0.71 & 0.76 & 0.73 \\
    Refute   & 0.96 & 0.99 & 0.97 & 0.96 & 0.99 & 0.97 & 0.98 & 0.98 & 0.98 \\
    \midrule
    Weighted Avg. & 0.76 & 0.76 & 0.75 & 0.76 & 0.76 & 0.76 & 0.76 & 0.76 & 0.76\\
    \bottomrule
 \end{tabular}
 }
\end{table}

\begin{table}
 \caption{5-way Classification Results of experiments with SBERT-QA + ER on val set}\label{tab:5wayresults_ex3}
 \small
 \resizebox{\columnwidth}{!}{
 \begin{tabular}{lSSSSSSSSSSSSS}
    \toprule
    \multirow{1}{*}{Categories} &
     \multicolumn{3}{|c}{SBERT-QA\_paragraph\_ER\_top5} &
     \multicolumn{3}{|c}{SBERT-QA\_sentence\_ER\_top5}  \\
     & {P} & {R} & {F1} & {P} & {R} & {F1} \\
     \midrule
    Support\_Multimodal & 0.80 & 0.77 & 0.78 & 0.79 & 0.83 & \textbf{0.81} \\
    Support\_Text  & 0.70 & 0.68 & 0.69 & 0.70 & 0.69 & \textbf{0.70} \\
    Insufficient\_Multimodal  & 0.66 & 0.72 & 0.69 & 0.71 & 0.72 & \textbf{0.73} \\
    Insufficient\_Text  & 0.76 & 0.72 & \textbf{0.74} & 0.74 & 0.72 & 0.73 \\
    Refute   & 0.96 & 1.00 & 0.98 & 0.99 & 0.98 & 0.98 \\
    \midrule
    Weighted Avg. & 0.78 & 0.78 & 0.78 & 0.79 & 0.79 & \textbf{0.79}\\
    \bottomrule
 \end{tabular}
 }
\end{table}

\subsection{Competition Result}
The final test set results and competition leaderboard are presented in Table \ref{tab:factify_leaderboard}. The results show that top 3 participating systems achieves similar performance and our system is ranked at 3rd place with a small margin (by 0.028) to the top performing system. Please refer to \cite{surya2023factifyoverview} for the competition details.

\begin{table}
 \caption{Factify Official Leaderboard}\label{tab:factify_leaderboard}
 \small
 \resizebox{\columnwidth}{!}{
 \begin{tabular}{lSSSSSSS}
    \toprule
     \multicolumn{1}{|c}{Rank} &
     \multicolumn{1}{|c}{Team} &
     \multicolumn{1}{|c}{Support\_Text} &
     \multicolumn{1}{|c}{Support\_Multi.} &
     \multicolumn{1}{|c}{Insufficient\_Text} &
     \multicolumn{1}{|c}{Insufficient\_Multi.} &
     \multicolumn{1}{|c}{Refute} &
     \multicolumn{1}{|c}{Final}
     \\
     \midrule
   1 & {Triple-Check} & \textbf{0.828} & \textbf{0.914} & 0.852 & \textbf{0.892} & \textbf{1.0} & \textbf{0.818} \\
   2 & INO       & 0.812 & 0.9  & \textbf{0.888} & 0.852  & 0.999 & 0.808 \\
   3 & {Logically} & 0.804 & 0.905 & 0.844 & 0.856 & 0.985 & 0.79 \\
   4 & {Zhang}  & 0.766 & 0.879 & 0.816 & 0.879 & 0.999 & 0.774 \\
   5 & {gzw} & 0.785 & 0.863 & 0.814 & 0.833 & 1.0 & 0.761 \\
   6 & {coco} & 0.773 & 0.865 & 0.815 & 0.83 & 1.0 & 0.757 \\
   7 & {Noir} & 0.771 & 0.873 & 0.785 & 0.816 & 	0.997 & 0.745 \\
   8 & {Yet} & 0.707 & 0.826 & 0.786 & 0.719 & 1.0 & 0.691 \\
   9 & {TeamX} & 0.582 & 0.709 & 0.537 & 0.556 & 0.698 & 0.456 \\
   \midrule
    - & {BASELINE} & 0.5 & 0.827 & 0.802 & 0.759 & 0.988 & 0.65 \\
    \bottomrule
 \end{tabular}}
\end{table}

\section{Conclusion}\label{sec:conclude}

In this research, we present our multimodal fact checking system that is submitted to the De-Factify 2023 competition. The system consists of various components, including a multimodal fact checking dataset, a QA-enhanced evidence passage retrieval component, and a Transformer-based cross-modal sequence-to-sequence veracity prediction model. Our findings from the De-Factify 2023 competition show that recent advances in pre-trained cross-modal models, such as CLIP, have strong zero-shot or few-shot capabilities and can be effectively transferred to a variety of downstream tasks, including multimodal fact checking. However, there is still a need for more effective techniques for multimodal modeling and explainability, particularly in regards to learning finer-grained cross-modal representations by jointly modeling intra- and inter-modality relationships and aligning vision regions with sentence words or entities. Additionally, more focus should be placed on real-world challenges that involve handling large amounts of textual and multimodal information from multiple sources and domains for claim verification. There is also a need for techniques that can effectively handle more complex and nuanced real-world scenarios, such as those involving sarcasm, irony, and misleading context. The difficulties in creating large and high-quality multimodal fact checking datasets that accurately reflect real-world scenarios (e.g., insufficient/leaked evidence), as identified in previous work \cite{glockner2022missing,gao2021log}, remain a significant challenge.








\bibliography{factify}

\begin{thebibliography}{44}
\expandafter\ifx\csname natexlab\endcsname\relax\def\natexlab#1{#1}\fi
\providecommand{\url}[1]{\texttt{#1}}
\providecommand{\href}[2]{#2}
\providecommand{\path}[1]{#1}
\providecommand{\DOIprefix}{doi:}
\providecommand{\ArXivprefix}{arXiv:}
\providecommand{\URLprefix}{URL: }
\providecommand{\Pubmedprefix}{pmid:}
\providecommand{\doi}[1]{\href{http://dx.doi.org/#1}{\path{#1}}}
\providecommand{\Pubmed}[1]{\href{pmid:#1}{\path{#1}}}
\providecommand{\bibinfo}[2]{#2}
\ifx\xfnm\relax \def\xfnm[#1]{\unskip,\space#1}\fi
\bibitem[{Wadden et~al.(2022)Wadden, Lo, Wang, Cohan, Beltagy, and
  Hajishirzi}]{wadden2022multivers}
\bibinfo{author}{D.~Wadden}, \bibinfo{author}{K.~Lo},
  \bibinfo{author}{L.~Wang}, \bibinfo{author}{A.~Cohan},
  \bibinfo{author}{I.~Beltagy}, \bibinfo{author}{H.~Hajishirzi},
\newblock \bibinfo{title}{Multivers: Improving scientific claim verification
  with weak supervision and full-document context},
\newblock in: \bibinfo{booktitle}{Findings of the Association for Computational
  Linguistics: NAACL 2022}, \bibinfo{year}{2022}, pp. \bibinfo{pages}{61--76}.
\bibitem[{Stammbach(2021)}]{stammbach2021evidence}
\bibinfo{author}{D.~Stammbach},
\newblock \bibinfo{title}{Evidence selection as a token-level prediction task},
\newblock in: \bibinfo{booktitle}{Proceedings of the Fourth Workshop on Fact
  Extraction and VERification (FEVER), EMNLP},
  \bibinfo{organization}{Association for Computational Linguistics (ACL)},
  \bibinfo{year}{2021}.
\bibitem[{Gao et~al.(2021)Gao, Hoffmann, Oikonomou, Kiskovski, and
  Bandhakavi}]{gao2021log}
\bibinfo{author}{J.~Gao}, \bibinfo{author}{H.-F. Hoffmann},
  \bibinfo{author}{S.~Oikonomou}, \bibinfo{author}{D.~Kiskovski},
  \bibinfo{author}{A.~Bandhakavi},
\newblock \bibinfo{title}{Logically at factify 2022: Multimodal fact
  verification},
\newblock \bibinfo{journal}{arXiv}  (\bibinfo{year}{2021}). \URLprefix
  \url{https://arxiv.org/abs/2112.09253}.
  \DOIprefix\doi{10.48550/ARXIV.2112.09253}.
\bibitem[{Zaheer et~al.(2020)Zaheer, Guruganesh, Dubey, Ainslie, Alberti,
  Ontanon, Pham, Ravula, Wang, Yang et~al.}]{zaheer2020big}
\bibinfo{author}{M.~Zaheer}, \bibinfo{author}{G.~Guruganesh},
  \bibinfo{author}{K.~A. Dubey}, \bibinfo{author}{J.~Ainslie},
  \bibinfo{author}{C.~Alberti}, \bibinfo{author}{S.~Ontanon},
  \bibinfo{author}{P.~Pham}, \bibinfo{author}{A.~Ravula},
  \bibinfo{author}{Q.~Wang}, \bibinfo{author}{L.~Yang}, et~al.,
\newblock \bibinfo{title}{Big bird: Transformers for longer sequences},
\newblock \bibinfo{journal}{Advances in Neural Information Processing Systems}
  \bibinfo{volume}{33} (\bibinfo{year}{2020}) \bibinfo{pages}{17283--17297}.
\bibitem[{Beltagy et~al.(2020)Beltagy, Peters, and
  Cohan}]{beltagy2020longformer}
\bibinfo{author}{I.~Beltagy}, \bibinfo{author}{M.~E. Peters},
  \bibinfo{author}{A.~Cohan},
\newblock \bibinfo{title}{Longformer: The long-document transformer},
\newblock \bibinfo{journal}{arXiv preprint arXiv:2004.05150}
  (\bibinfo{year}{2020}).
\bibitem[{Ma et~al.(2022)Ma, Zhou, Kong, He, Gui, Neubig, May, and
  Zettlemoyer}]{ma2022mega}
\bibinfo{author}{X.~Ma}, \bibinfo{author}{C.~Zhou}, \bibinfo{author}{X.~Kong},
  \bibinfo{author}{J.~He}, \bibinfo{author}{L.~Gui},
  \bibinfo{author}{G.~Neubig}, \bibinfo{author}{J.~May},
  \bibinfo{author}{L.~Zettlemoyer},
\newblock \bibinfo{title}{Mega: Moving average equipped gated attention},
\newblock \bibinfo{journal}{arXiv preprint arXiv:2209.10655}
  (\bibinfo{year}{2022}).
\bibitem[{Smith et~al.(2022)Smith, Warrington, and
  Linderman}]{smith2022simplified}
\bibinfo{author}{J.~T. Smith}, \bibinfo{author}{A.~Warrington},
  \bibinfo{author}{S.~W. Linderman},
\newblock \bibinfo{title}{Simplified state space layers for sequence modeling},
\newblock \bibinfo{journal}{arXiv preprint arXiv:2208.04933}
  (\bibinfo{year}{2022}).
\bibitem[{Zhao et~al.(2021)Zhao, Bao, Wang, Zhou, Wu, He, and
  Zhou}]{zhao2021ror}
\bibinfo{author}{J.~Zhao}, \bibinfo{author}{J.~Bao}, \bibinfo{author}{Y.~Wang},
  \bibinfo{author}{Y.~Zhou}, \bibinfo{author}{Y.~Wu}, \bibinfo{author}{X.~He},
  \bibinfo{author}{B.~Zhou},
\newblock \bibinfo{title}{Ror: Read-over-read for long document machine reading
  comprehension},
\newblock \bibinfo{journal}{arXiv preprint arXiv:2109.04780}
  (\bibinfo{year}{2021}).
\bibitem[{Nakano et~al.(2021)Nakano, Hilton, Balaji, Wu, Ouyang, Kim, Hesse,
  Jain, Kosaraju, Saunders et~al.}]{nakano2021webgpt}
\bibinfo{author}{R.~Nakano}, \bibinfo{author}{J.~Hilton},
  \bibinfo{author}{S.~Balaji}, \bibinfo{author}{J.~Wu},
  \bibinfo{author}{L.~Ouyang}, \bibinfo{author}{C.~Kim},
  \bibinfo{author}{C.~Hesse}, \bibinfo{author}{S.~Jain},
  \bibinfo{author}{V.~Kosaraju}, \bibinfo{author}{W.~Saunders}, et~al.,
\newblock \bibinfo{title}{Webgpt: Browser-assisted question-answering with
  human feedback},
\newblock \bibinfo{journal}{arXiv preprint arXiv:2112.09332}
  (\bibinfo{year}{2021}).
\bibitem[{Wadden et~al.(2020)Wadden, Lin, Lo, Wang, van Zuylen, Cohan, and
  Hajishirzi}]{wadden2020fact}
\bibinfo{author}{D.~Wadden}, \bibinfo{author}{S.~Lin}, \bibinfo{author}{K.~Lo},
  \bibinfo{author}{L.~L. Wang}, \bibinfo{author}{M.~van Zuylen},
  \bibinfo{author}{A.~Cohan}, \bibinfo{author}{H.~Hajishirzi},
\newblock \bibinfo{title}{Fact or fiction: Verifying scientific claims},
\newblock \bibinfo{journal}{arXiv preprint arXiv:2004.14974}
  (\bibinfo{year}{2020}).
\bibitem[{Thorne et~al.(2018)Thorne, Vlachos, Christodoulopoulos, and
  Mittal}]{thorne-etal-2018-fever}
\bibinfo{author}{J.~Thorne}, \bibinfo{author}{A.~Vlachos},
  \bibinfo{author}{C.~Christodoulopoulos}, \bibinfo{author}{A.~Mittal},
\newblock \bibinfo{title}{Fever: a large-scale dataset for fact extraction and
  verification},
\newblock in: \bibinfo{booktitle}{Proceedings of the 2018 Conference of the
  North American Chapter of the Association for Computational Linguistics:
  Human Language Technologies, Volume 1 (Long Papers)},
  \bibinfo{publisher}{Association for Computational Linguistics},
  \bibinfo{address}{New Orleans, Louisiana}, \bibinfo{year}{2018}, pp.
  \bibinfo{pages}{809--819}. \URLprefix
  \url{https://aclanthology.org/N18-1074}.
  \DOIprefix\doi{10.18653/v1/N18-1074}.
\bibitem[{Aly et~al.(2021)Aly, Guo, Schlichtkrull, Thorne, Vlachos,
  Christodoulopoulos, Cocarascu, and Mittal}]{aly2021feverous}
\bibinfo{author}{R.~Aly}, \bibinfo{author}{Z.~Guo},
  \bibinfo{author}{M.~Schlichtkrull}, \bibinfo{author}{J.~Thorne},
  \bibinfo{author}{A.~Vlachos}, \bibinfo{author}{C.~Christodoulopoulos},
  \bibinfo{author}{O.~Cocarascu}, \bibinfo{author}{A.~Mittal},
\newblock \bibinfo{title}{Feverous: Fact extraction and verification over
  unstructured and structured information},
\newblock \bibinfo{journal}{arXiv preprint arXiv:2106.05707}
  (\bibinfo{year}{2021}).
\bibitem[{Floridi and Chiriatti(2020)}]{floridi2020gpt}
\bibinfo{author}{L.~Floridi}, \bibinfo{author}{M.~Chiriatti},
\newblock \bibinfo{title}{Gpt-3: Its nature, scope, limits, and consequences},
\newblock \bibinfo{journal}{Minds and Machines} \bibinfo{volume}{30}
  (\bibinfo{year}{2020}) \bibinfo{pages}{681--694}.
\bibitem[{Maynez et~al.(2020)Maynez, Narayan, Bohnet, and
  McDonald}]{maynez2020faithfulness}
\bibinfo{author}{J.~Maynez}, \bibinfo{author}{S.~Narayan},
  \bibinfo{author}{B.~Bohnet}, \bibinfo{author}{R.~McDonald},
\newblock \bibinfo{title}{On faithfulness and factuality in abstractive
  summarization},
\newblock \bibinfo{journal}{arXiv preprint arXiv:2005.00661}
  (\bibinfo{year}{2020}).
\bibitem[{Lazaridou et~al.(2022)Lazaridou, Gribovskaya, Stokowiec, and
  Grigorev}]{lazaridou2022internet}
\bibinfo{author}{A.~Lazaridou}, \bibinfo{author}{E.~Gribovskaya},
  \bibinfo{author}{W.~Stokowiec}, \bibinfo{author}{N.~Grigorev},
\newblock \bibinfo{title}{Internet-augmented language models through few-shot
  prompting for open-domain question answering},
\newblock \bibinfo{journal}{arXiv preprint arXiv:2203.05115}
  (\bibinfo{year}{2022}).
\bibitem[{Wei et~al.(2021)Wei, Bosma, Zhao, Guu, Yu, Lester, Du, Dai, and
  Le}]{wei2021finetuned}
\bibinfo{author}{J.~Wei}, \bibinfo{author}{M.~Bosma}, \bibinfo{author}{V.~Y.
  Zhao}, \bibinfo{author}{K.~Guu}, \bibinfo{author}{A.~W. Yu},
  \bibinfo{author}{B.~Lester}, \bibinfo{author}{N.~Du}, \bibinfo{author}{A.~M.
  Dai}, \bibinfo{author}{Q.~V. Le},
\newblock \bibinfo{title}{Finetuned language models are zero-shot learners},
\newblock \bibinfo{journal}{arXiv preprint arXiv:2109.01652}
  (\bibinfo{year}{2021}).
\bibitem[{Reimers and Gurevych(2019)}]{reimers2019sentence}
\bibinfo{author}{N.~Reimers}, \bibinfo{author}{I.~Gurevych},
\newblock \bibinfo{title}{Sentence-bert: Sentence embeddings using siamese
  bert-networks},
\newblock \bibinfo{journal}{arXiv preprint arXiv:1908.10084}
  (\bibinfo{year}{2019}).
\bibitem[{Devlin et~al.(2018)Devlin, Chang, Lee, and
  Toutanova}]{devlin2018bert}
\bibinfo{author}{J.~Devlin}, \bibinfo{author}{M.-W. Chang},
  \bibinfo{author}{K.~Lee}, \bibinfo{author}{K.~Toutanova},
\newblock \bibinfo{title}{Bert: Pre-training of deep bidirectional transformers
  for language understanding},
\newblock \bibinfo{journal}{arXiv preprint arXiv:1810.04805}
  (\bibinfo{year}{2018}).
\bibitem[{Saakyan et~al.(2021)Saakyan, Chakrabarty, and
  Muresan}]{saakyan2021covid}
\bibinfo{author}{A.~Saakyan}, \bibinfo{author}{T.~Chakrabarty},
  \bibinfo{author}{S.~Muresan},
\newblock \bibinfo{title}{Covid-fact: Fact extraction and verification of
  real-world claims on covid-19 pandemic},
\newblock \bibinfo{journal}{arXiv preprint arXiv:2106.03794}
  (\bibinfo{year}{2021}).
\bibitem[{Yao et~al.(2022)Yao, Shah, Sun, Cho, and Huang}]{yao2022end}
\bibinfo{author}{B.~M. Yao}, \bibinfo{author}{A.~Shah},
  \bibinfo{author}{L.~Sun}, \bibinfo{author}{J.-H. Cho},
  \bibinfo{author}{L.~Huang},
\newblock \bibinfo{title}{End-to-end multimodal fact-checking and explanation
  generation: A challenging dataset and models},
\newblock \bibinfo{journal}{arXiv preprint arXiv:2205.12487}
  (\bibinfo{year}{2022}).
\bibitem[{Barr{\'o}n-Cede{\~n}o et~al.(2020)Barr{\'o}n-Cede{\~n}o, Elsayed,
  Nakov, Da~San~Martino, Hasanain, Suwaileh, Haouari, Babulkov, Hamdan, Nikolov
  et~al.}]{barron2020overview}
\bibinfo{author}{A.~Barr{\'o}n-Cede{\~n}o}, \bibinfo{author}{T.~Elsayed},
  \bibinfo{author}{P.~Nakov}, \bibinfo{author}{G.~Da~San~Martino},
  \bibinfo{author}{M.~Hasanain}, \bibinfo{author}{R.~Suwaileh},
  \bibinfo{author}{F.~Haouari}, \bibinfo{author}{N.~Babulkov},
  \bibinfo{author}{B.~Hamdan}, \bibinfo{author}{A.~Nikolov}, et~al.,
\newblock \bibinfo{title}{Overview of checkthat! 2020: Automatic identification
  and verification of claims in social media},
\newblock in: \bibinfo{booktitle}{International Conference of the
  Cross-Language Evaluation Forum for European Languages},
  \bibinfo{organization}{Springer}, \bibinfo{year}{2020}, pp.
  \bibinfo{pages}{215--236}.
\bibitem[{Lee et~al.(2018)Lee, Wu, and Fung}]{lee2018improving}
\bibinfo{author}{N.~Lee}, \bibinfo{author}{C.-S. Wu},
  \bibinfo{author}{P.~Fung},
\newblock \bibinfo{title}{Improving large-scale fact-checking using
  decomposable attention models and lexical tagging},
\newblock in: \bibinfo{booktitle}{Proceedings of the 2018 Conference on
  Empirical Methods in Natural Language Processing}, \bibinfo{year}{2018}, pp.
  \bibinfo{pages}{1133--1138}.
\bibitem[{Schuster et~al.(2021)Schuster, Fisch, and Barzilay}]{schuster2021get}
\bibinfo{author}{T.~Schuster}, \bibinfo{author}{A.~Fisch},
  \bibinfo{author}{R.~Barzilay},
\newblock \bibinfo{title}{Get your vitamin c! robust fact verification with
  contrastive evidence},
\newblock \bibinfo{journal}{arXiv preprint arXiv:2103.08541}
  (\bibinfo{year}{2021}).
\bibitem[{Sun et~al.(2019)Sun, Myers, Vondrick, Murphy, and
  Schmid}]{sun2019video}
\bibinfo{author}{C.~Sun}, \bibinfo{author}{A.~Myers},
  \bibinfo{author}{C.~Vondrick}, \bibinfo{author}{K.~Murphy},
  \bibinfo{author}{C.~Schmid},
\newblock \bibinfo{title}{Videobert: A joint model for video and language
  representation learning},
\newblock \bibinfo{journal}{arXiv}  (\bibinfo{year}{2019}). \URLprefix
  \url{https://arxiv.org/abs/1904.01766}.
  \DOIprefix\doi{10.48550/ARXIV.1904.01766}.
\bibitem[{Li et~al.(2019)Li, Yatskar, Yin, Hsieh, and Chang}]{li2019visual}
\bibinfo{author}{L.~H. Li}, \bibinfo{author}{M.~Yatskar},
  \bibinfo{author}{D.~Yin}, \bibinfo{author}{C.-J. Hsieh},
  \bibinfo{author}{K.-W. Chang},
\newblock \bibinfo{title}{Visualbert: A simple and performant baseline for
  vision and language},
\newblock \bibinfo{journal}{arXiv}  (\bibinfo{year}{2019}). \URLprefix
  \url{https://arxiv.org/abs/1908.03557}.
  \DOIprefix\doi{10.48550/ARXIV.1908.03557}.
\bibitem[{Chen et~al.(2019)Chen, Li, Yu, Kholy, Ahmed, Gan, Cheng, and
  Liu}]{chen2019uniter}
\bibinfo{author}{Y.-C. Chen}, \bibinfo{author}{L.~Li}, \bibinfo{author}{L.~Yu},
  \bibinfo{author}{A.~E. Kholy}, \bibinfo{author}{F.~Ahmed},
  \bibinfo{author}{Z.~Gan}, \bibinfo{author}{Y.~Cheng},
  \bibinfo{author}{J.~Liu},
\newblock \bibinfo{title}{Uniter: Universal image-text representation
  learning},
\newblock \bibinfo{journal}{arXiv}  (\bibinfo{year}{2019}). \URLprefix
  \url{https://arxiv.org/abs/1909.11740}.
  \DOIprefix\doi{10.48550/ARXIV.1909.11740}.
\bibitem[{Radford et~al.(2021)Radford, Kim, Hallacy, Ramesh, Goh, Agarwal,
  Sastry, Askell, Mishkin, Clark et~al.}]{radford2021learning}
\bibinfo{author}{A.~Radford}, \bibinfo{author}{J.~W. Kim},
  \bibinfo{author}{C.~Hallacy}, \bibinfo{author}{A.~Ramesh},
  \bibinfo{author}{G.~Goh}, \bibinfo{author}{S.~Agarwal},
  \bibinfo{author}{G.~Sastry}, \bibinfo{author}{A.~Askell},
  \bibinfo{author}{P.~Mishkin}, \bibinfo{author}{J.~Clark}, et~al.,
\newblock \bibinfo{title}{Learning transferable visual models from natural
  language supervision},
\newblock in: \bibinfo{booktitle}{International Conference on Machine
  Learning}, \bibinfo{organization}{PMLR}, \bibinfo{year}{2021}, pp.
  \bibinfo{pages}{8748--8763}.
\bibitem[{Gao et~al.(2021)Gao, Geng, Zhang, Ma, Fang, Zhang, Li, and
  Qiao}]{gao2021clip}
\bibinfo{author}{P.~Gao}, \bibinfo{author}{S.~Geng},
  \bibinfo{author}{R.~Zhang}, \bibinfo{author}{T.~Ma},
  \bibinfo{author}{R.~Fang}, \bibinfo{author}{Y.~Zhang},
  \bibinfo{author}{H.~Li}, \bibinfo{author}{Y.~Qiao},
\newblock \bibinfo{title}{Clip-adapter: Better vision-language models with
  feature adapters},
\newblock \bibinfo{journal}{arXiv preprint arXiv:2110.04544}
  (\bibinfo{year}{2021}).
\bibitem[{Guo et~al.(2022)Guo, Zhang, Qiu, Ma, Miao, He, and
  Cui}]{guo2022calip}
\bibinfo{author}{Z.~Guo}, \bibinfo{author}{R.~Zhang}, \bibinfo{author}{L.~Qiu},
  \bibinfo{author}{X.~Ma}, \bibinfo{author}{X.~Miao}, \bibinfo{author}{X.~He},
  \bibinfo{author}{B.~Cui},
\newblock \bibinfo{title}{Calip: Zero-shot enhancement of clip with
  parameter-free attention},
\newblock \bibinfo{journal}{arXiv preprint arXiv:2209.14169}
  (\bibinfo{year}{2022}).
\bibitem[{Jiang et~al.(2022)Jiang, He, Xu, and Wang}]{jiang2022comclip}
\bibinfo{author}{K.~Jiang}, \bibinfo{author}{X.~He}, \bibinfo{author}{R.~Xu},
  \bibinfo{author}{X.~E. Wang},
\newblock \bibinfo{title}{Comclip: Training-free compositional image and text
  matching},
\newblock \bibinfo{journal}{arXiv preprint arXiv:2211.13854}
  (\bibinfo{year}{2022}).
\bibitem[{Wang and Peng(2022)}]{yao2022fact}
\bibinfo{author}{W.-Y. Wang}, \bibinfo{author}{W.-C. Peng},
\newblock \bibinfo{title}{Team yao at factify 2022: Utilizing pre-trained
  models and co-attention networks for multi-modal fact verification},
\newblock \bibinfo{journal}{arXiv}  (\bibinfo{year}{2022}). \URLprefix
  \url{https://arxiv.org/abs/2201.11664}.
  \DOIprefix\doi{10.48550/ARXIV.2201.11664}.
\bibitem[{Messina et~al.(2021)Messina, Amato, Esuli, Falchi, Gennaro, and
  Marchand-Maillet}]{messina2021fine}
\bibinfo{author}{N.~Messina}, \bibinfo{author}{G.~Amato},
  \bibinfo{author}{A.~Esuli}, \bibinfo{author}{F.~Falchi},
  \bibinfo{author}{C.~Gennaro}, \bibinfo{author}{S.~Marchand-Maillet},
\newblock \bibinfo{title}{Fine-grained visual textual alignment for cross-modal
  retrieval using transformer encoders},
\newblock \bibinfo{journal}{ACM Transactions on Multimedia Computing,
  Communications, and Applications (TOMM)} \bibinfo{volume}{17}
  (\bibinfo{year}{2021}) \bibinfo{pages}{1--23}.
\bibitem[{Suryavardan et~al.(2023)Suryavardan, Mishra, Patwa, Chakraborty,
  Rani, Reganti, Chadha, Das, Sheth, Chinnakotla, Ekbal, and
  Kumar}]{surya2023factify2}
\bibinfo{author}{S.~Suryavardan}, \bibinfo{author}{S.~Mishra},
  \bibinfo{author}{P.~Patwa}, \bibinfo{author}{M.~Chakraborty},
  \bibinfo{author}{A.~Rani}, \bibinfo{author}{A.~Reganti},
  \bibinfo{author}{A.~Chadha}, \bibinfo{author}{A.~Das},
  \bibinfo{author}{A.~Sheth}, \bibinfo{author}{M.~Chinnakotla},
  \bibinfo{author}{A.~Ekbal}, \bibinfo{author}{S.~Kumar},
\newblock \bibinfo{title}{Factify 2: A multimodal fake news and satire news
  dataset},
\newblock in: \bibinfo{booktitle}{proceedings of defactify 2: second workshop
  on Multimodal Fact-Checking and Hate Speech Detection},
  \bibinfo{publisher}{CEUR}, \bibinfo{year}{2023}.
\bibitem[{Thorne et~al.(2018)Thorne, Vlachos, Cocarascu, Christodoulopoulos,
  and Mittal}]{thorne2018fact}
\bibinfo{author}{J.~Thorne}, \bibinfo{author}{A.~Vlachos},
  \bibinfo{author}{O.~Cocarascu}, \bibinfo{author}{C.~Christodoulopoulos},
  \bibinfo{author}{A.~Mittal},
\newblock \bibinfo{title}{The fact extraction and verification (fever) shared
  task},
\newblock \bibinfo{journal}{arXiv preprint arXiv:1811.10971}
  (\bibinfo{year}{2018}).
\bibitem[{Thorne et~al.(2019)Thorne, Vlachos, Cocarascu, Christodoulopoulos,
  and Mittal}]{thorne2019fever2}
\bibinfo{author}{J.~Thorne}, \bibinfo{author}{A.~Vlachos},
  \bibinfo{author}{O.~Cocarascu}, \bibinfo{author}{C.~Christodoulopoulos},
  \bibinfo{author}{A.~Mittal},
\newblock \bibinfo{title}{The fever2. 0 shared task},
\newblock in: \bibinfo{booktitle}{Proceedings of the Second Workshop on Fact
  Extraction and VERification (FEVER)}, \bibinfo{year}{2019}, pp.
  \bibinfo{pages}{1--6}.
\bibitem[{Aly et~al.(2021)Aly, Guo, Schlichtkrull, Thorne, Vlachos,
  Christodoulopoulos, Cocarascu, and Mittal}]{aly2021fact}
\bibinfo{author}{R.~Aly}, \bibinfo{author}{Z.~Guo}, \bibinfo{author}{M.~S.
  Schlichtkrull}, \bibinfo{author}{J.~Thorne}, \bibinfo{author}{A.~Vlachos},
  \bibinfo{author}{C.~Christodoulopoulos}, \bibinfo{author}{O.~Cocarascu},
  \bibinfo{author}{A.~Mittal},
\newblock \bibinfo{title}{The fact extraction and verification over
  unstructured and structured information (feverous) shared task},
\newblock in: \bibinfo{booktitle}{Proceedings of the Fourth Workshop on Fact
  Extraction and VERification (FEVER)}, \bibinfo{year}{2021}, pp.
  \bibinfo{pages}{1--13}.
\bibitem[{Wadden and Lo(2021)}]{wadden2021overview}
\bibinfo{author}{D.~Wadden}, \bibinfo{author}{K.~Lo},
\newblock \bibinfo{title}{Overview and insights from the sciver shared task on
  scientific claim verification},
\newblock \bibinfo{journal}{arXiv preprint arXiv:2107.08188}
  (\bibinfo{year}{2021}).
\bibitem[{Jiang et~al.(2021)Jiang, Pradeep, and Lin}]{jiang2021exploring}
\bibinfo{author}{K.~Jiang}, \bibinfo{author}{R.~Pradeep},
  \bibinfo{author}{J.~Lin},
\newblock \bibinfo{title}{Exploring listwise evidence reasoning with t5 for
  fact verification},
\newblock in: \bibinfo{booktitle}{Proceedings of the 59th Annual Meeting of the
  Association for Computational Linguistics and the 11th International Joint
  Conference on Natural Language Processing (Volume 2: Short Papers)},
  \bibinfo{year}{2021}, pp. \bibinfo{pages}{402--410}.
\bibitem[{Song et~al.(2020)Song, Tan, Qin, Lu, and Liu}]{song2020mpnet}
\bibinfo{author}{K.~Song}, \bibinfo{author}{X.~Tan}, \bibinfo{author}{T.~Qin},
  \bibinfo{author}{J.~Lu}, \bibinfo{author}{T.-Y. Liu},
\newblock \bibinfo{title}{Mpnet: Masked and permuted pre-training for language
  understanding},
\newblock \bibinfo{journal}{Advances in Neural Information Processing Systems}
  \bibinfo{volume}{33} (\bibinfo{year}{2020}) \bibinfo{pages}{16857--16867}.
\bibitem[{Mikolov et~al.(2013)Mikolov, Sutskever, Chen, Corrado, and
  Dean}]{mikolov2013distributed}
\bibinfo{author}{T.~Mikolov}, \bibinfo{author}{I.~Sutskever},
  \bibinfo{author}{K.~Chen}, \bibinfo{author}{G.~S. Corrado},
  \bibinfo{author}{J.~Dean},
\newblock \bibinfo{title}{Distributed representations of words and phrases and
  their compositionality},
\newblock \bibinfo{journal}{Advances in neural information processing systems}
  \bibinfo{volume}{26} (\bibinfo{year}{2013}).
\bibitem[{Vaswani et~al.(2017)Vaswani, Shazeer, Parmar, Uszkoreit, Jones,
  Gomez, Kaiser, and Polosukhin}]{vaswani2017att}
\bibinfo{author}{A.~Vaswani}, \bibinfo{author}{N.~Shazeer},
  \bibinfo{author}{N.~Parmar}, \bibinfo{author}{J.~Uszkoreit},
  \bibinfo{author}{L.~Jones}, \bibinfo{author}{A.~N. Gomez},
  \bibinfo{author}{L.~Kaiser}, \bibinfo{author}{I.~Polosukhin},
\newblock \bibinfo{title}{Attention is all you need},
\newblock \bibinfo{journal}{arXiv}  (\bibinfo{year}{2017}). \URLprefix
  \url{https://arxiv.org/abs/1706.03762}.
  \DOIprefix\doi{10.48550/ARXIV.1706.03762}.
\bibitem[{Loshchilov and Hutter(2017)}]{losh2017adamw}
\bibinfo{author}{I.~Loshchilov}, \bibinfo{author}{F.~Hutter},
\newblock \bibinfo{title}{Decoupled weight decay regularization}
  (\bibinfo{year}{2017}). \URLprefix \url{https://arxiv.org/abs/1711.05101}.
  \DOIprefix\doi{10.48550/ARXIV.1711.05101}.
\bibitem[{Suryavardan et~al.(2023)Suryavardan, Mishra, Chakraborty, Patwa,
  Rani, Chadha, Reganti, Das, Sheth, Chinnakotla, Ekbal, and
  Kumar}]{surya2023factifyoverview}
\bibinfo{author}{S.~Suryavardan}, \bibinfo{author}{S.~Mishra},
  \bibinfo{author}{M.~Chakraborty}, \bibinfo{author}{P.~Patwa},
  \bibinfo{author}{A.~Rani}, \bibinfo{author}{A.~Chadha},
  \bibinfo{author}{A.~Reganti}, \bibinfo{author}{A.~Das},
  \bibinfo{author}{A.~Sheth}, \bibinfo{author}{M.~Chinnakotla},
  \bibinfo{author}{A.~Ekbal}, \bibinfo{author}{S.~Kumar},
\newblock \bibinfo{title}{Findings of factify 2: multimodal fake news
  detection},
\newblock in: \bibinfo{booktitle}{proceedings of defactify 2: second workshop
  on Multimodal Fact-Checking and Hate Speech Detection},
  \bibinfo{publisher}{CEUR}, \bibinfo{year}{2023}.
\bibitem[{Glockner et~al.(2022)Glockner, Hou, and
  Gurevych}]{glockner2022missing}
\bibinfo{author}{M.~Glockner}, \bibinfo{author}{Y.~Hou},
  \bibinfo{author}{I.~Gurevych},
\newblock \bibinfo{title}{Missing counter-evidence renders nlp fact-checking
  unrealistic for misinformation},
\newblock \bibinfo{journal}{arXiv preprint arXiv:2210.13865}
  (\bibinfo{year}{2022}).

\end{thebibliography}




\end{document}